\documentclass[10pt,twocolumn,letterpaper]{article}

\usepackage{iccv,times,amsmath,amssymb,graphicx,subfigure}
\usepackage{paralist,booktabs,multirow,booktabs,multirow}
\usepackage[pagebackref=true,breaklinks=true,letterpaper=true,colorlinks,bookmarks=false]{hyperref}

\iccvfinalcopy 

\ificcvfinal\fi

\newcommand{\tabincell}[2]{\begin{tabular}{@{}#1@{}}#2\end{tabular}}

\makeatletter\renewcommand\paragraph{\@startsection{paragraph}{4}{\z@}
	{.5em \@plus1ex \@minus.2ex}{-.5em}{\normalfont\normalsize\bfseries}}\makeatother

\begin{document}
	
	\title{Learning Efficient Video Representation with Video Shuffle Networks \vspace{-3mm}}
	\author{Pingchuan Ma\textsuperscript{1}\thanks{Equal contribution.} \qquad Yao Zhou\textsuperscript{2}\footnotemark[1] \qquad Yu Lu\textsuperscript{3} \qquad Wei Zhang\textsuperscript{3}\\[2mm]
		\textsuperscript{1} MIT \qquad \textsuperscript{2} Tencent YouTu Lab\qquad \textsuperscript{3} SenseTime Research\\
		{\tt\small pika7ma@gmail.com \qquad yoosan.zhou@gmail.com \quad \{luyu,wayne.zhang\}@sensetime.com}
	}\maketitle

	\begin{abstract}
		3D CNN shows its strong ability in learning spatiotemporal representation in recent video recognition tasks. However, inflating 2D convolution to 3D inevitably introduces additional computational costs, making it cumbersome in practical deployment. We consider whether there is a way to equip the conventional 2D convolution with temporal vision no requiring expanding its kernel. To this end, we propose the video shuffle, a parameter-free plug-in component that efficiently reallocates the inputs of 2D convolution so that its receptive field can be extended to the temporal dimension. In practical, video shuffle firstly divides each frame feature into multiple groups and then aggregate the grouped features via temporal shuffle operation. This allows the following 2D convolution aggregate the global spatiotemporal features. The proposed video shuffle can be flexibly inserted into popular 2D CNNs, forming the Video Shuffle Networks (VSN). With a simple yet efficient implementation, VSN performs surprisingly well on temporal modeling benchmarks. In experiments, VSN not only gains non-trivial improvements on Kinetics and Moments in Time, but also achieves state-of-the-art performance on Something-Something-V1, Something-Something-V2 datasets.
		
	\end{abstract}
	
	\section{Introduction}
	
	\begin{figure}[t]
	\centering
	\includegraphics[scale=.46]{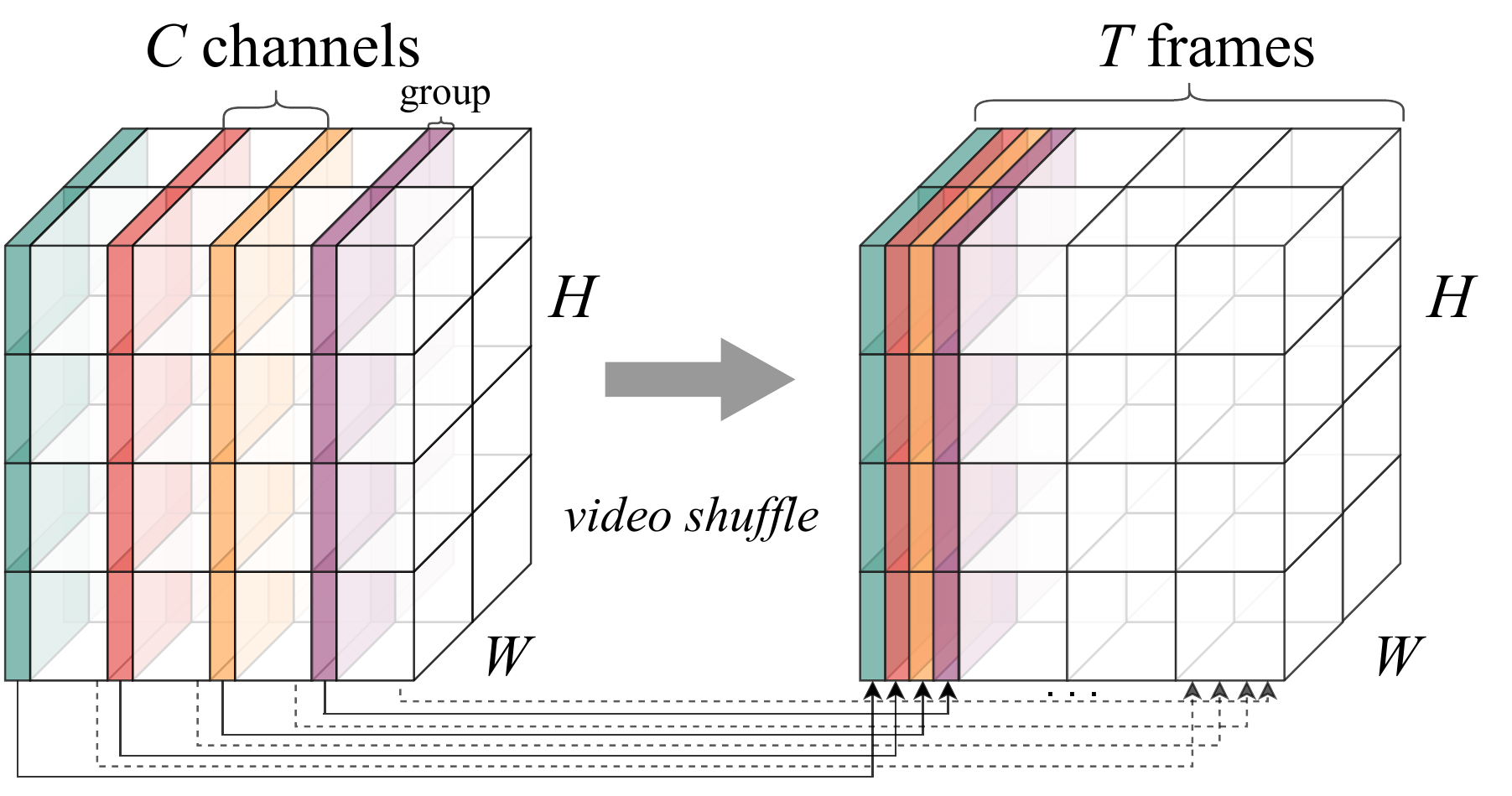} \caption{The proposed \textit{video shuffle} first divides channels of each frame feature into several groups with equal sizes, and then stacks the grouped features at same index along temporal dimension into a new frame feature. Through video shuffle, spatial information is exchanged across all frames.}
	\label{fig-model}\vspace{-3mm}
    \end{figure}
    
	End-to-end learning methods have achieved great improvements over previous hand-crafted features~\cite{klaser2008spatio,laptev2005space,Wang2013ActionRW}, and become the mainstream in video recognition area.
    The design of video recognition models enjoys great benefits of the prior art of still image recognition models. On one hand, many works utilize successful 2D CNNs, such as Inception~\cite{szegedy2015going} and ResNet~\cite{he2016deep} architectures, to extract spatial features of individual frames and then perform temporal aggregation using pooling strategies~\cite{karpathy2014large,girdhar2017attentional,wang2016temporal}, feature encoding functions~\cite{girdhar2017actionvlad,miech2017learnable,zhou2018temporal}, recurrent neural networks~\cite{donahue2015long,yue2015beyond,li2017temporal}, and even optical flow-guided methods~\cite{fan2018end,sun2018optical,repflow2019}. These approaches incorporate a learnable module into 2D CNNs that captures temporal dependency and motion information. On the other hand, some works directly inflate 2D CNNs into 3D CNNs by replacing $k\times k$ convolution filters with $k\times k\times k$~\cite{carreira2017quo,tran2015learning}, and then add non-local block~\cite{wang2018non} to grasp long-range temporal dependency or separate $k\times k\times k$ kernel into $1\times k\times k$ and $k\times 1 \times 1$ kernels~\cite{qiu2017learning,xie2017rethinking,tran2018closer} to reduce computational costs. The expanded temporal filters in 3D CNNs thus can conveniently model the temporal information from videos. Although these improved 3D CNNs show their effectiveness for action recognition, they usually introduce additional computational cost. This may limit the usage of 3D CNNs in real-world applications requiring low latency.
    
    In conventional 2D video models, each frame is independently fed into a 2D CNN to extract feature and then a temporal pooling function aggregates all frame features to video level. We consider the probability of performing temporal modeling by reallocating the inputs of 2D convolution, instead of introducing the temporal integrating methods. To this end, we equip 2D convolution with spatio-temporal receptive field by employing the recent group~\cite{krizhevsky2012imagenet,xie2017rethinking} and shuffle operations~\cite{zhang2018shufflenet}. Specifically, we propose video shuffle, an efficient and generic plug-in component for modeling temporal dependency in 2D CNNs with zero cost. As shown in Figure~\ref{fig-model}, video shuffle first divides channels of each frame into several groups with equal size, and then aggregates all of grouped features with same group index into a new frame feature. The reallocated frame feature contains spatial information of all frames and therefore the following 2D convolutions can conveniently learn both spatial and temporal representation. Video shuffle is superiority efficient since there are no additional parameters and FLOPs (addition or multiplication) introduced. The computation time of the proposed video shuffle comes only from the data movement in memory, which hardly affects the inference latency.
    
	Video shuffle can be easily incorporated into 2D video models. In this work, we adopt temporal segment networks (TSN)~\cite{wang2016temporal} as our basic model, and take ResNet-50 and ResNet-101~\cite{he2016deep} as the backbones. In implementation, we plug video shuffle and its inverse operation, which restores the original spatial representation of each frame, into ResNet, before and after 2D convolutions inside residual blocks respectively. To demonstrate the effectiveness of the proposed VSN, we conduct experiments on several popular video action recognition datasets, including large-scale Kinetics and Moments in Time as well as the temporal-sensitive Something-Somethings. In experiments, VSN outperforms its counterpart on all datasets at the cost of zero parameters and zero FLOPs. Moreover, VSN surpass it by a large margin and further achieves state-of-the-art performance on the challenging Something-Something-V1, Something-Something-V2 datasets. 
	
    \section{Related Work}
    \subsection{Video Recognition Models}
    The conventional 2D CNNs learn video representation using 2D CNNs as spatial features extractor for frames and then performing temporal aggregation over frame features. In \cite{karpathy2014large}, they made use of 2D CNNs to extract features from individual frames and then integrated temporal features into a fixed-size video representation using various fusion methods. Many works focused on designing temporal aggregation methods to improve the recognition accuracy. Pooling approaches~\cite{miech2017learnable,girdhar2017attentional,wang2016temporal}, feature encoding functions~\cite{girdhar2017actionvlad,zhou2018temporal} and recurrent neural networks~\cite{donahue2015long,yue2015beyond,li2017temporal} were usually preformed on high-level features, while optical flow-guided methods computed motion information on low and middle-level features~\cite{fan2018end,sun2018optical}. Two-stream framework introduced by ~\cite{simonyan2014two} is a widely-used approach to capture the motion information. It fused deep features extracted from optical flows and traditional features computed from RGB inputs.
    
    On the other hand, a video can be viewed as a cube stacked of many images. That is to say, a 3D convolution can process video directly. Previous works demonstrate that 3D CNNs ~\cite{tran2015learning} can straightforward learn the spatio-temporal features. In order to take benefits of the successful 2D CNNs and ImageNet pretraining, Carreira and Zisserman~\cite{carreira2017quo} introduces the Inflated 3D ConvNets (I3D) based on the Inception architecture~\cite{szegedy2015going,ioffe2015batch} and show its superior performance on a large human action recognition benchmark~\cite{kay2017kinetics}. Meanwhile, several 3D variants are proposed~\cite{qiu2017learning,xie2017rethinking,tran2018closer,hara2018can}. Qiu $\etal$~\cite{qiu2017learning} introduces a Pseudo-3D ResNet architecture. Tran $\etal$~\cite{tran2018closer} presents the R(2+1)D model based on ResNet. Xie $\etal$~\cite{xie2017rethinking} investigates where ``deflating'' 3D convolution are more suitable and then presents the separable-3D model built upon I3D. These mixed 2D and 3D networks are constructed by replacing $k\times k\times k$ filter to $1\times k\times k$ followed by a $k\times 1\times 1$ filter. Additionally, non-local neural network~\cite{wang2018non} and its improved version~\cite{wang2018videos}, trajectory convolution~\cite{zhao2018trajectory}, energy-based models~\cite{wang2018appearance} are also introduced.
    
    In parallel, some works focused on efficient model design in video understanding. The most related approaches were ECO~\cite{zolfaghari2018eco} and TSM~\cite{lin2018temporal}. ECO~\cite{zolfaghari2018eco} employed a 3D-net stacking on the 2D feature extractors to model the temporal relationships, and they further proposed an online video understanding algorithm for fast video inference. Although ECO achieves a good runtime-accuracy trade-off, it still increased the computational costs compared to 2D CNNs. TSM~\cite{lin2018temporal} introduced a zero-cost {temporal shift module} which \textit{shift}s part of channels along temporal dimension by $\pm 1$ to fuse temporal information. TSM not only had $2.7\times$ fewer FLOPs than ECO family but also achieved finer recognition performance, especially on temporal-sensitive datasets.
    
    \subsection{Group and Shuffle Operation}
    The idea of splitting channels into several groups was first presented in AlexNet~\cite{krizhevsky2012imagenet} for distributing the model over two GPUs to handle the memory issue, and then widely used for designing tiny and efficient network architectures~\cite{howard2017mobilenets}. In ResNeXt~\cite{xie2017aggregated}, they further developed the group convolution to reduce the number of parameters and computational complexity by dividing input channels into several groups, then performing regular convolution on each group and concatenating all group results as the outputs. Experiments demonstrated that group convolution can lead to performance improvement on image recognition task. But when stacking group convolution multiple layers, the outputs from a certain channel were only derived from a small fraction of input channels. To address this weakness, ShuffleNet~\cite{zhang2018shufflenet} utilized the channel shuffle operation, by which the resulting channels of each group were collected from all input groups, to enable information interaction across groups. It not only greatly reduced computational costs but also maintained accuracy. Besides, Zhang $etal$~\cite{zhang2017interleaved} proposed an interleaved group convolutions which consists of a primary group convolution for handling spatial correlation and a secondary group convolution for blending the channels and show its effectiveness. A special case of group convolutions was channel-wise convolution where the number of groups is equal to the number of channels, this is also very similar to the separable convolution~\cite{chollet2017xception,howard2017mobilenets}. The basic idea of group and shuffle operations are adopted in this work.
    
	\begin{figure}[t]
	\centering
	\includegraphics[scale=.50]{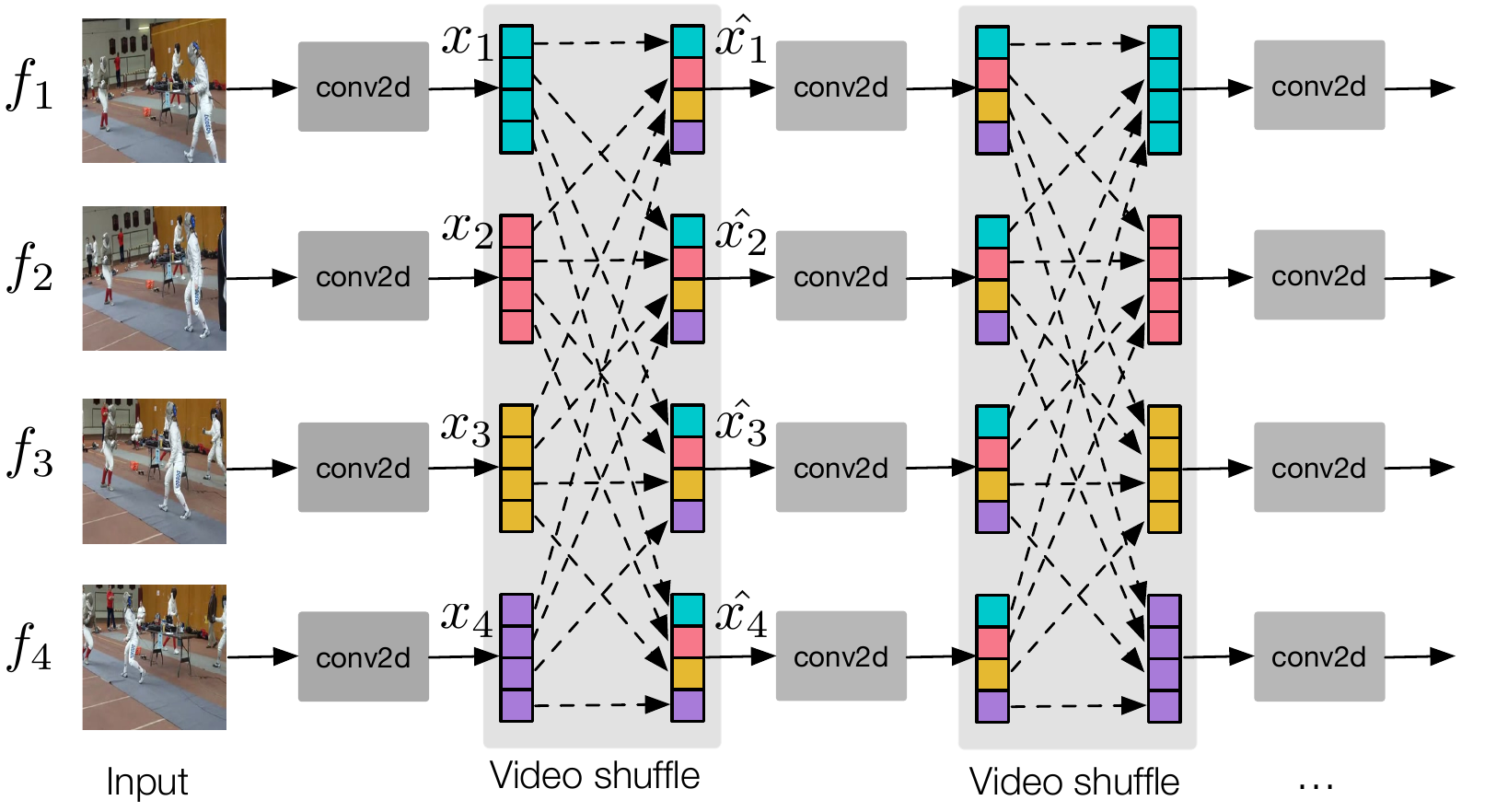}
	\caption{Graphical representation of \textit{video shuffle networks}. Isolated frame features $x_i$ are aggregated into video-level feature $\hat{x_i}$ via \textit{video shuffle} operations. In this manner, instead of focusing on an instant moment, subsequent 2D convolutions are endowed with non-local perception and thus learn video representation more effectively and efficiently.}
	\label{fig-vsn}\vspace{-3mm}
    \end{figure}

    \section{Video Shuffle Networks}
    \label{vsn}
    In this section, we first introduce the design criteria of video shuffle, and show how to incorporate it into the building block of ResNet. Then, we present the overall network architecture of VSN, followed by implementation details.

    \subsection{The Design of Video Shuffle}
    \label{vsu}
    The design motivation of video shuffle lies in the fact that though recent 3D CNNs have improved recognition performance, they could hardly be deployed into real-world video recognition systems due to its heavy computational cost. The conventional 2D CNNs enjoy low latency, but they learn spatial feature from isolated frames without temporal modeling, leading to accuracy gap against state-of-the-arts models. Namely, there is an accuracy-speed trade-off in video recognition models. To address it, we propose to equip 2D CNNs with temporal receptive field by reallocating the inputs of 2D convolution. In order to enable 2D convolution learn both spatial and temporal feature without modifying its structure, there are two prerequisites: 1) the input should contain spatial representation from all frames and 2) its input size should not be changed. Group operation divides input channels into several groups to make each contain partial representation and shuffle operation further facilitates information exchanging across different groups. These features make them perfect choices in this work.

    Figure~\ref{fig-vsn} shows a graphical representation of the proposed video shuffle. A video with $T$ frame features is shown as an example and each one of them is a $[C, H, W]$ tensor extracted by 2D convolutions, where $C$ indicates the channel size, $H$ and $W$ are spatial dimensions. For each frame feature, we first divide channels into several groups with equal sizes. In this work, number of groups is heuristically set to number of frames $T$, in consequence, channel size of each grouped feature equals $C/T$. In this way, each grouped feature with shape of $[C/T, H, W]$ contains a part of spatial feature. We aggregate all of grouped features with same group index into a new frame feature by temporal shuffle operation, which allows spatial features exchanging across different frames. As illustrated in Figure~\ref{fig-model} right: the reallocated feature at the first frame is a stack of all first grouped features in Figure~\ref{fig-model} left (before video shuffle applied).

    Denoting a feature at $i$-th frame as $x_{i}$, video shuffle transforms the original $x_{i}$ to a new feature $\hat{x_{i}}$ by the following equation:
    \begin{equation}
    \hat{x_{i}} = [x_{1}^{(i-1)\eta:i\eta}, x_{2}^{(i-1)\eta:i\eta} , \ldots , x_{T}^{(i-1)\eta:i\eta}],
    \end{equation}
    where $1 \leqslant i \leqslant T$, $\eta$ is the channel size of each group and $\eta = C/T$ in this setting, the symbol $:$ denotes the tensor slicing operation along channel dimension. The index $i$ plays the role of both frame and group index. For instance, $x_{1}^{0:\eta}$ indicates the first grouped feature at the first frame (masked {green} in Figure~\ref{fig-model} left). As a result, the new frame feature $\hat{x_{i}}$ contains the spatial information of all sequential frames and further serves as the inputs of the following 2D convolutions. The proposed video shuffle has three advantages: first, it allows spatial features interacting across different frames; second, the following 2D convolutions in 2D CNNs can handily perform both spatial and temporal modeling; and third, video shuffle is easy to implement via data movement in memory, not introducing additional parameters or theoretical FLOPs at all.

    \subsection{Video Shuffle in Residual Block}
    \label{vsb}
    Since we have obtained a parameter-free video shuffle that is able to model temporal information cooperating with 2D convolutions, we consider inserting it into conventional 2D CNNs. In this work, we mainly study on the ResNet architectures. As a result, we attempt to plug video shuffle and its reverse operation, which restores the original spatial representation for each frame, into the primary building block of ResNet. We investigated two positions to insert video shuffle units and obtain two variants: the \textit{headtail (residual) block} and \textit{compact (residual) block}.

    \begin{figure}[htbp]
    	\centering
    	\subfigure[headtail]{\label{fig-headtail} \includegraphics[scale=0.34]{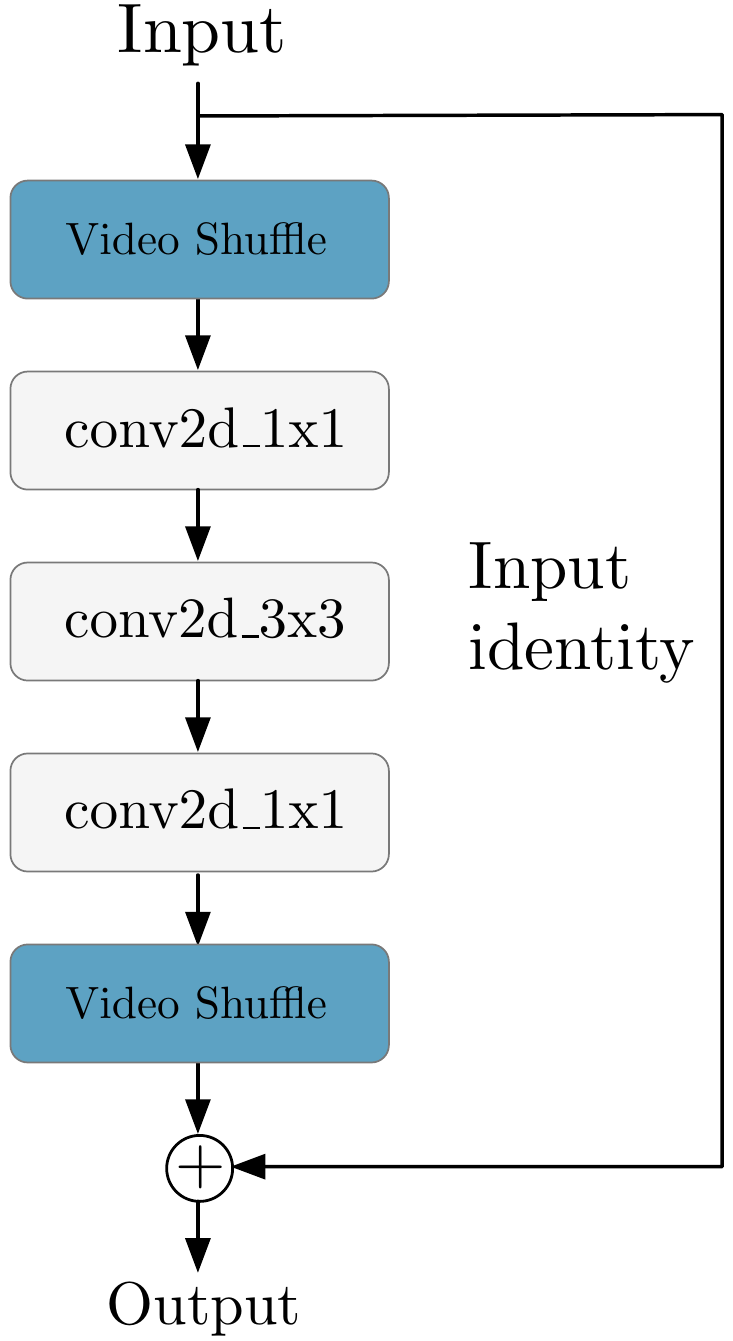}}
    	\hspace{40px}
    	\subfigure[compact]{\label{fig-compact} \includegraphics[scale=0.34]{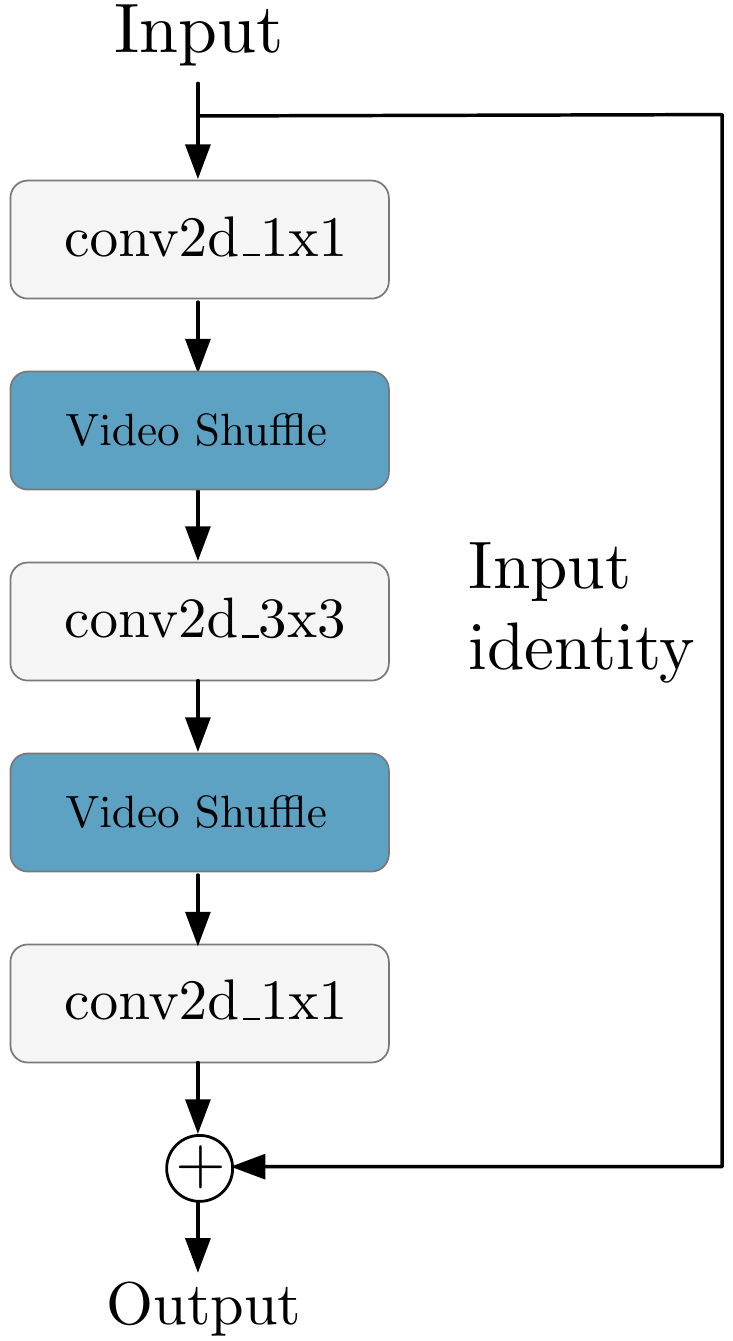}}
    	\caption{A headtail block places video shuffle at the head and tail of a residual block while a compact block performs video shuffle before and after conv2d\_$3$x$3$ immediately.}
    	\label{fig-block}
    \end{figure}

    We first consider placing video shuffle at the head and tail of a residual block, namely \textit{headtail block}. Before any convolutional layer, the inputs go though the first video shuffle directly, and each new frame is consequently composed with partial spatial features from all sampled frames. The following convolutions in bottleneck block end-to-end learn both spatio-temporal feature consequently. After them, to guide subsequent convolutions focusing on spatial reasoning, we restore the original spatial feature of each frame by inserting an inverse video shuffle. As for \textit{compact block}, video shuffle units are similarly configured with a ``paired'' setting but compactly performed before and after conv2d\_3x3 (shown in Figure~\ref{fig-block}).

    Given that a bottleneck performs compression on channel dimension where the spatial and temporal information blend, we argue that a weakness of headtail residual block could be an information loss along with dimensional reduction. We further empirically verify this assumption in experiments (ablation studies). Results demonstrate that compact block is stronger than headtail in temporal modeling. Unless specified, we always use the compact block in following experiments.

    \subsection{Network Architectures}
    \label{secnet}
    
    \begin{table}[ht]
    	\begin{center}
    		\scriptsize
    		\begin{tabular}{c|c|c|c}
    			\hline
    			layer name & ResNet-50 & ResNet-101 & output size \\
    			\hline
    			$\textnormal{conv}_1$ & \multicolumn{2}{c|}{$7\times 7$, stride 2} & 64$\times$8$\times$112$\times$112 \\
    			\hline
    			$\textnormal{pool}_1$ & \multicolumn{2}{c|}{$3\times 3$ max, stride 2} & 64$\times$8$\times$56$\times$56 \\
    			\hline
    			$\textnormal{res}_2$ & $\tabincell{c}{standard block $\times$2 \\compact block $\times$1}$ & $\tabincell{c}{standard block $\times$2 \\compact block $\times$1} $ & 256$\times$8$\times$56$\times$ 56 \\
    			\hline
    			$\textnormal{res}_3$ & $\tabincell{c}{standard block $\times$3 \\compact block $\times$1}$ & $\tabincell{c}{standard block $\times$3 \\compact block $\times$1} $ & 512$\times$8$\times$28$\times$ 28 \\
    			\hline
    			$\textnormal{res}_4$ & $\tabincell{c}{standard block $\times$5 \\compact block $\times$1}$ & $\tabincell{c}{standard block $\times$22 \\compact block $\times$1} $ & 1024$\times$8$\times$14$\times$ 14 \\
    			\hline
    			$\textnormal{res}_5$ & $\tabincell{c}{standard block $\times$2 \\compact block $\times$1}$ & $\tabincell{c}{standard block $\times$2 \\compact block $\times$1} $ & 2048$\times$8$\times$7$\times$ 7 \\
    			\hline
    			\multicolumn{3}{c|}{global average pooling} & 2048$\times$1$\times$1$\times$1 \\
    			\hline
    		\end{tabular}
    	\end{center}
    	\caption{The proposed VSN-ResNet-50 and VSN-ResNet-101 architectures for video recognition.}
    	\label{tab-net}\vspace{-3mm}
    \end{table}

    Table~\ref{tab-net} presents VSN with ResNet-50 and ResNet-101 backbone for video recognition. In this work, we have not added or modified any convolution or pooling layer. Instead of incorporating video shuffle into all building blocks, we heuristically insert it into the last building block in each ResNet layer. For example, in $\textit{res}_2$ of ResNet-50, we replace the third block with the compact video shuffle and retain the first two blocks. That is, there are \textbf{totally four blocks} equipped with video shuffle in overall ResNet architectures, and more related ablation studies are presented in experiments.

    In TSM~\cite{lin2018temporal}, they insert the temporal shift module into the residual block and show its effectiveness in video recognition. We argue that TSM only allows temporal information to be interchanged between neighbor frames, which models temporal information locally. As a result, it fails to take advantages of non-local details in long range. Compared with local-field TSM, video shuffle broadens its horizon to all frames and models temporal dependency in a global version. As TSM is orthogonal to video shuffle, we further combine them together. In terms of implementation, we use their \textit{residual temporal shift module} with zero padding to replace the building blocks which have not be incorporated with video shuffle e.g. the first two blocks in $\textit{res}_2$ of ResNet-50. Experiments are conducted to show the superiority of such combination in temporal modeling. In practice, we replace the last block of $\textit{res}_2$, $\textit{res}_3$, $\textit{res}_4$ and $\textit{res}_5$ with our compact block and add temporal shift module to the other residual blocks. We denote our video model as \textit{VSN-ResNet-L} (\textit{VSN-RL} for simplicity) if the backbone is \textit{ResNet-L}, where $L$ indicates the number of layers, e.g. \textit{VSN-ResNet-50 (VSN-R50) and VSN-ResNet-101 (VSN-R101)}.

    \begin{table*}[ht]
    	\begin{center}
    		\small
    		\begin{tabular}{lccccccc}
    			\toprule
    			Model  & \#Frame & \#Params & FLOPs & Sth-V1 val & Sth-V1 test & Sth-V2 val & Sth-V2 test\\ 
    			\midrule
    			TSN~\cite{wang2016temporal}  & 8 & 10.7M & 16G & 19.5 & - & 33.4 & -\\
    			TRN-Multiscale~\cite{zhou2018temporal}  & 8 & 18.3M & 16G & 34.4 & 33.6 & 48.8 & 50.9\\
    			Two-stream TRN~\cite{zhou2018temporal}  & 8+8 & 36.6M & - & 42.0 & 40.7 & 55.5 & 56.2\\
    			ECO~\cite{zolfaghari2018eco} & 16 & 47.5M & 64G & 41.4 & - & - & - \\
    			$\textnormal{ECO}_{En}\_{Lite}$~\cite{zolfaghari2018eco} & 92 & 150M & 267G & 46.4 & 42.3 & - & - \\ 
    			$\textnormal{ECO}_{En}\_{Lite}_\textnormal{R+F}$~\cite{zolfaghari2018eco} & 92+92 & 300M & - & 49.5 & 43.9 & - & - \\
    			I3D~\cite{wang2018non} & 64 & 28.0M & 306G & 41.6 & - & - & - \\
    			NL-I3D~\cite{wang2018non} & 64 & 35.3M & 335G & 44.4 & - & - & -\\
    			NL-I3D + GCN~\cite{wang2018videos} & 64 & 62.2M & 605G & 46.1 & 45.0 & - & - \\
    			TrajectoryNet~\cite{zhao2018trajectory} & 32 & 33.3M & - & 47.8 & - & - & - \\
    			TSM~\cite{lin2018temporal} & 8 & 24.3M & 33G & 43.4 & - & 59.1 & - \\
    			TSM~\cite{lin2018temporal} & 16 & 24.3M & 65G & 44.8 & - & 59.4 & 60.4 \\
    			Two-stream TSM~\cite{lin2018temporal} & 16+16 & 48.6M & - & 50.2 & 47.0 & 64.0 & 64.3\\
    			\midrule
    			$\textnormal{VSN-R50}_\textnormal{RGB}$ & 8 & 24.3M & 33G & 46.6 & - & 60.6 & - \\
    			$\textnormal{VSN-R50}_\textnormal{Flow}$ & 8 & 24.3M & 33G & 38.8 & - & 53.8 & -\\
    			Two-stream VSN-R50 & 8+8 & 48.6M & - & 51.6 & - & 65.3 & - \\
    			\midrule
    			$\textnormal{VSN-R101}_\textnormal{RGB}$ & 8 & 42.9M & 63G & \textbf{47.8} & - & \textbf{61.6} & - \\
    			$\textnormal{VSN-R101}_\textnormal{Flow}$ & 8 & 42.9M & 63G & 41.4 & - & 56.7 & - \\
    			$\textnormal{VSN-R\{50+101\}}_\textnormal{RGB}$ & 8 & 67.2M & 96G & {49.2} & 46.8 & 63.2 & 64.6 \\
    			Two-stream VSN-R101 & 8+8 & 85.8M & - & \textbf{52.7} & \textbf{49.9} & \textbf{65.8} & \textbf{66.1} \\
    			\bottomrule
    		\end{tabular}
    	\end{center}
    	\caption{Comparison of the proposed VSN with the previous state-of-the-art models on Sth-V1 and Sth-V2 (Top-1 accuracy).}
    	\label{tab-sth}\vspace{-3mm}
    \end{table*}

    \subsection{Implementation Details}
    In this work, all models are implemented in PyTorch. ResNet architectures and ImageNet pre-trained models are derived from \textit{torchvision} package.

    \paragraph{Training.}
    We sample 8 frames from an entire video using the sparse segment-based sampling~\cite{wang2016temporal}. For data augmentation, our implementation follows the practice in~\cite{wang2016temporal} to alleviate negative effects of overfitting. The images are first resized with shorter side to $256$ and then applied by corner cropping and scale-jittering. We also apply random left-right flipping consistently for all videos except actions are horizontal-order-sensitive in Something-Something. e.g. \textit{Pushing something from left to right}. Finally, the cropped images are resized to $224\times 224$ pixels for network training. We distribute totally 64 videos into 8 TITANXP GPUs and each GPU has 8 videos in a mini-batch. We adopt SGD with momentum as optimizer and set its initial learning rate to 0.01. We utilize both the multi-step learning rate decaying and cosine learning rate schedule~\cite{loshchilov2016sgdr} with warm-up depending on dataset. The momentum value, weight decay and dropout rate are set to 0.9, 5e-4 and 0.8 respectively. We freeze all batch normalization except the first convolution layer.

    \paragraph{Inference.}
    In the inference phase, TSN~\cite{wang2016temporal} takes the average predictions of 25$\times$10 crops as the video prediction. I3D and S3D~\cite{xie2017aggregated} densely sample all frames and take center crops for evaluation. In this work, we take the same pre-processing as non-local neural network~\cite{wang2018non}, which performs spatially fully-convolutional inference on videos whose shorter side is re-scaled to 256. For temporal domain, we also sample total 8 frames during evaluation.

    \section{Experiments}
    In this paper, extensive experiments are performed on four popular and challenging video recognition benchmarks. We first introduce these experimental benchmarks and then show that the proposed VSN can not only perform very well on Kinetics, but also achieve state-of-the-art performance on Something-Something-V1, Something-Something-V2 and Moments in Time datasets.

    \subsection{Datasets}
    We conduct experiments on various video datasets with great diversity, whose sources range from YouTube to crowdsourcing videos and durations range from three seconds to tens of seconds, covering human daily activities, human actions to sports and events.

    \noindent \textbf{Kinetics}~\cite{kay2017kinetics} is a large human action recognition dataset, which contains around 240k training videos and 20k validation videos, involving 400 human cation classes.

    \noindent \textbf{Moments in Time (Moments)}~\cite{monfort2019moments} includes a collection of 1 million trimmed 3s-video clips, corresponding to 339 dynamic event categories.

    \noindent \textbf{Something-Something-V1 (Sth-V1)}~\cite{goyal2017something} is a temporal-sensitive dataset, containing 108,499 videos. The total 174 categories are basic actions with objects.

    \noindent \textbf{Something-Something-V2 (Sth-V2)}~\cite{goyal2017something} increases its number of videos to 220,847 and further improves the annotation quality and pixel resolution.

    \begin{table}[ht]
    	\begin{center}
    		\scriptsize
    		\begin{tabular}{llcc}
    			\toprule
    			Model  & Backbone   & Top-1 & Top-5 \\ 
    			\midrule
    			TSN~\cite{wang2016temporal} & Inception-V3 & 71.5 & 90.2 \\
    			Attention-Cluster~\cite{long2018attention} & Inc-Res-v2 & 75.0 & 91.9 \\
    			NL-C2D~\cite{wang2018non}\ & ResNet-101  & 75.1 & 91.6 \\
    			CPNet~\cite{liu2019learning} & ResNet-101 & 75.3 & 92.4 \\
    			\midrule
    			I3D~\cite{carreira2017quo} & Inception & 72.1 & 90.3 \\
    			R(2+1)D~\cite{tran2018closer} & ResNet-34  & 74.3 & 91.4 \\
    			S3D-G~\cite{xie2017rethinking} & Inception  & 74.7 & 93.4 \\
    			CoST~\cite{li2019collaborative} & ResNet-101  & 77.5 & 93.2 \\
    			NL-I3D~\cite{wang2018non} & ResNet-101  & 77.7 & 93.3 \\
    			SlowFast+NL~\cite{feichtenhofer2018slowfast} & ResNet-101 & \textbf{79.8} & \textbf{93.9} \\
    			\midrule
    			VSN-R50 & ResNet-50  & 73.5 & 91.3 \\
    			VSN-R101 & ResNet-101  &  75.4 & 92.2 \\
    			Two stream VSN-R101 & ResNet-101 & 77.6 & 93.7 \\
    			\bottomrule
    		\end{tabular}
    	\end{center}
    	\caption{Comparison of VSN and the previous state-of-the-art models on the validation set of Kinetics.}
    	\label{tab-kinetics}\vspace{-3mm}
    \end{table}

    \subsection{Results on Something-Something}

    \paragraph{Something-Something-V1.} We first show a comparison of the performance between VSN and previous state-of-the-art methods in Table~\ref{tab-sth}, on both validation and test set of Something-Something-V1. The top-1 accuracy as well as the statistics of computational costs are reported.

    Previous results are presented in the first group. \cite{zhou2018temporal} found that TSN fails to reason temporal relation and thus proposed the temporal relation networks (TRN) to learn temporal dependencies between video frames at multiple time scales. They show that TRN-multiscale can improve TSN by 14.7\% and fusing optical flow gives another 7.6\% improvement. In~\cite{zolfaghari2018eco}, they introduced the efficient video understanding model ECO by leveraging the 3D-net stacking on 2D features. Their best single model achieved an accuracy of 41.4\%. Some works attempted to use pure 3D models. Both I3D~\cite{carreira2017quo} and its improved version NL-I3D~\cite{wang2018non} obtained good performance, but their computational costs (FLOPs) are too huge to deploy. Furthermore, to explicitly model relationships between humans and objects, NL-I3D+GCN~\cite{wang2018videos} used a object detector to extract region proposals and composed these regions from different frames by the graph convolution network. Although NL-I3D+GCN achieved a very competitive accuracy of 46.1\%, the introduced computational cost is non-negligible. TrajectoryNet~\cite{zhao2018trajectory} allows visual features to be aggregated along motion paths by a trajectory convolution, achieving a higher accuracy of 47.8\%. The recent temporal shift module (TSM) achieved 43.4\% when taking 8 RGB frames as inputs. As number of frames was increased to 16, it gained another 1.4\% boosts. The previous state-of-the-art performance was held by $\textnormal{TSM}_{\textnormal{RGB+Flow}}$, which fused 16-frames RGB model with another optical flow stream.

    Our results are presented at the last two groups. Taking 8 RGB frames as inputs, our VSN-R50 achieves 46.6\% accuracy, outperforming TSN and TSM by 27.1\% and 3.2\% respectively. This demonstrates that video shuffle performs outstanding for temporal modeling. When fused with optical flow stream, two-stream VSN-R50 achieves 51.6\% accuracy, which is 1.4\% higher than two-stream TSM. Going deeper with network architecture from ResNet-50 to ResNet-101 gives notable 2.2\% improvements. The best single model VSN-R101 gets an accuracy of 47.8\%. Our ensemble model, which averages the predictions of VSN-R101 and VSN-R50, achieves top-1 accuracy of 49.2\%. Furthermore, the two-stream $\textnormal{VSN-R101}$ gets a new state-of-the-art 52.7\% performance on Something-Something-V1 dataset. We also submit test predictions to the evaluation server and report test results. The trend of improvement is basically consistent with the validation set and the best performance 49.9\% is held by our two-stream VSN-R101.

    \paragraph{Something-Something-V2.} It distinguishes from previous Sth-V1 dataset with increased training examples, better annotation quality and higher video resolution. The comparison results of VSN with the state-of-the-art methods also listed in Table~\ref{tab-sth}. The previous models have been introduced in the above experiments. Similar to Sth-V1 dataset, our models are able to outperform both 2D and 3D counterparts. VSN-R50 and VSN-R101 achieve 60.6\% and 61.6\% respectively. With VSN going deeper, the improvement gains constantly climb higher. VSN-R101 outperforms TSM by 2.5\%. In accord with our expectation, the ensemble models ($\textnormal{VSN-R\{50+101\}}_\textnormal{RGB}$) show big advantages over the ones fed with single modality. Evaluated on the validation and test set, our two-stream VSN-R101 establishes the new state-of-the-art on both sets.

    \subsection{Results on Kinetics and Moments in Time}

    \begin{table}[ht]
    	\begin{center}
    		\scriptsize
    		\begin{tabular}{llcc}
    			\toprule
    			Model  & Backbone & Top-1 & Top-5 \\ 
    			\midrule
    			TSN~\cite{wang2016temporal} & BN-Inception & 24.11 & - \\
    			Two-stream TSN~\cite{wang2016temporal} & BN-Inception & 25.32 & 50.10 \\
    			TRN~\cite{zhou2018temporal} & BN-Inception & 25.97 & - \\
    			Two-stream TRN~\cite{zhou2018temporal} & BN-Inception & 28.27 & 53.87 \\
    			ResNet50-ImageNet~\cite{monfort2019moments} & ResNet-50 & 27.16 & 51.68 \\
    			Two-stream I3D~\cite{monfort2019moments} & Inception-V1 &  29.51 &
    			56.06 \\
    			\midrule 
    			$\textnormal{CoST}_{8F}$~\cite{li2019collaborative} & ResNet-50 & 30.10 & 57.20  \\
    			$\textnormal{CoST}_{8F}$~\cite{li2019collaborative} & ResNet-101 & 31.50 & 57.90  \\
    			$\textnormal{CoST}_{32F}$~\cite{li2019collaborative} & ResNet-101 & 32.40 & 60.00  \\
    			\midrule
    			$\textnormal{VSN-R50}_{8F}$ & ResNet-50 & 31.74 & 58.66 \\
    			$\textnormal{VSN-R101}_{8F}$ & ResNet-101 & \textbf{32.65} &  \textbf{61.47} \\
    			\bottomrule
    		\end{tabular}
    	\end{center}
    	\caption{Comparison with state-of-the-arts on Moments in Time dataset. Our models are trained only on RGB inputs.}
    	\label{tab-moments}\vspace{-3mm}
    \end{table}

    \paragraph{Kinetics and Moments} Our VSN models are also evaluated on both Kinetics and Moments in Time, featuring huge size and tough task. Table~\ref{tab-kinetics} and Table~\ref{tab-moments} compares our VSN-R50 and VSN-R101 models against the previous state-of-the-art models on Kinetics and Moments respectively. First, it is observed that VSN outperforms TSN by a considerable margin, which verifies the effectiveness of the proposed video shuffle component. Second, compared with 2D attention-based models, such as Attention-Cluster and NL-C2D listed in Table~\ref{tab-kinetics}, our VSN-R101 achieves a very close performance. It demonstrates video shuffle indeed can act as a non-local feature integrator. Third, VSN-R101 can outperform 3D variants, such as I3D, R(2+1)D and S3D-G. Even comparing to huge 3D counterparts, our two-stream VSN-R101 is also competitive. Although both 2D and 3D models are not well-performed on the challenging Moments, VSN can outperform these counterparts. VSN-R101 beats all other models and stands as a new state-of-the-art.

    \begin{table}
    	\begin{center}
    	\small
    		\begin{tabular}{lccc}
    			\toprule
    			Model  & \#Flow & Kinetics & Sth-V1 \\ 
    			\midrule
    			ResNet-50~\cite{wang2016temporal} & $8\times 1$ & 47.5 & 27.0 \\
    			ResNet-50~\cite{wang2016temporal} & $8\times 5$ & 54.8 & 34.3 \\
    			ResNet-101~\cite{wang2016temporal} & $8\times 1$ & 49.7 & 29.2 \\
    			ResNet-101~\cite{wang2016temporal} & $8\times 5$ &  56.5 & 34.3 \\
    			\midrule
    			VSN-R50 & $8\times 1$ & 53.0 & 33.7 \\
    			VSN-R50 & $8\times 5$ &  56.7 & 37.5 \\
    			VSN-R101 & $8\times 1$ &  56.0 & 36.1 \\
    			VSN-R101 & $8\times 5$ &  60.1 & 41.4 \\
    			\bottomrule
    		\end{tabular}
    	\end{center}
    	\caption{Comparison of VSN models against TSN counterparts on Kinetics and Sth-V1, trained on optical flow inputs.}
    	\label{tab-flow}\vspace{-3mm}
    \end{table}

    \subsection{Generalize to Optical Flow}
    We also verify whether VSN can generalize to optical flow. For these experiments, we follow the standard setup as described in~\cite{simonyan2014two} and extract optical flow with the TV-L1 algorithm~\cite{perez2013tv}. All models are trained on the Kinetics and Sth-V1 and report the top-1 accuracy. We sample 8 segments in training optical flows like RGB. In~\cite{simonyan2014two,wang2016temporal}, they stack 5 or 10 consecutive optical flows for capturing the long-term temporal dependency in videos. We consider that VSN have ability to learn long-range temporal dependency and verify it by training models using only 1 optical flow in a segment.

    The results are shown in Table~\ref{tab-flow}. The first group presents the performance of TSN baseline with different backbones while the second one presents ours. Trained with $8\times1$ flow as inputs on Kinetics and Sth-V1, our VSN-R50 outperforms its counterpart by 5.5\% and 6.7\% respectively. By increasing the number of inputting flows from 1 to 5, both baseline and our model yield considerable gains. Furthermore, our VSN-R50 trained on $8\times 1$ optical flows is able to achieve performance close to the TSN ResNet-50 with $8\times 5$ flows, whose inputs are $5\times$ more than ours. Going deeper with VSN, the improvement grows considerably. VSN-R101 outperforms VSN-R50 models by around 3\%-4\% accuracy.

    \begin{table}
    \begin{center}
    \scriptsize
    \begin{tabular}{llccc}
    \toprule
    Model    & Latency & Throughput & Sth Acc(\%) \\
    \midrule
    I3D~\cite{carreira2017quo}        & 165.3ms & 6.1vps & 41.6 \\
    TSN-R50~\cite{wang2016temporal}      & 15.8ms & 80.8vps & 20.2  \\
    TSN-R101~\cite{wang2016temporal}     & 26.7ms & 48.8vps & 22.7    \\
    $\textnormal{ECO}_{16F}$ (Zolfaghari et.al 2018)   & 30.6ms & 45.6vps & 41.4 \\
    $\textnormal{TSM}_{8F}$~\cite{lin2018temporal}    & 17.4ms & 77.4vps & 43.4 \\
    \midrule
    VSN-R50   & 16.5ms & 79.5vps    & 44.5   \\
    VSN-R101    & 28.7ms & 47.2vps   & 46.5  \\
    \bottomrule
    \end{tabular}
    \end{center}
    \caption{Comparison in latency of VSN against the others.}
    \label{tab-latency}
    \end{table}

    \subsection{Inference Latency}
    To measure the latency and throughput, we perform inference on one NVIDIA Tesla P100 GPU and use the average value of 500 times batch inference with batch size of 16. Following~\cite{lin2018temporal}, we provide the speed of VSN-R50 and VSN-101. The \textit{vps} indicates the videos per second. It is clearly observed from Table~\ref{tab-latency} that our VSN models act superior not only by high accuracy but also by low latency and high throughput. Compared to I3D, VSN gets 13$\times$ speedup along with higher accuracy. It is also illustrated that video shuffle can hardly hurt the runtime speed: VSN has almost the same latency and throughput as TSN, but it brings 20\%+ improvement.

    \subsection{Ablation Studies}
    \label{sec-abs}

    \begin{table}
    	\begin{center}
    	\scriptsize
    		\begin{tabular}{llll}
    			\toprule
    			Backbone & Block  & Kinetics & Sth-V1 \\ 
    			\hline
    			\multirow{3}{*}{ResNet-50} & baseline & 71.5 & 20.2 \\
    			& headtail & $\text{72.1}^\text{+0.6}$ & $\text{43.2}^\text{+23.0}$ \\
    			& compact & $\textbf{73.5}^\textbf{+2.0}$ & $\textbf{46.6}^\textbf{+26.4}$ \\
    			\hline
    			\multirow{3}{*}{ResNet-101} & baseline & 72.8 & 22.7 \\
    			& headtail & $\text{74.0}^\text{+1.2}$ & $\text{44.5}^\text{+21.8}$ \\
    			& compact & $\textbf{75.4}^\textbf{+2.6}$ & $\textbf{47.8}^\textbf{+25.1}$ \\
    			\bottomrule
    		\end{tabular}
    	\end{center}
    	\caption{Comparison with different residual blocks.}
    	\label{tab-block}
    \end{table}

    \paragraph{Which residual block is better for temporal modeling?} Table~\ref{tab-block} shows top-1 accuracies of compact and headtail residual block both on Kinetics and Sth-V1. In this setting, we train our models with RGB inputs and replace all last blocks at different ResNet layers with video shuffle blocks. Although both of two variants outperform the baseline, compact residual block clearly outperforms headtail counterpart, no matter testing on temporal-sensitive dataset or using backbones network with different depth.

    \begin{table}
    	\begin{center}
    	\scriptsize
    		\begin{tabular}{cll}
    			\toprule
    			\#Blocks & Kinetics  & Sth-V1  \\ 
    			\midrule
    			0  & 71.5 & 20.2 \\
    			1  &  $\text{72.2}^\text{+0.7}$ & $\text{40.7}^\text{+20.5}$ \\
    			2  &  $\text{73.0}^\text{+1.5}$ & $\text{42.2}^\text{+22.0}$  \\
    			3  &  $\text{72.8}^\text{+1.3}$ & $\text{43.9}^\text{+23.7}$ \\
    			4  & $\textbf{73.5}^\textbf{+2.0}$ & $\textbf{46.6}^\textbf{+26.4}$ \\
    			\toprule
    		\end{tabular}
    	\end{center}
    	\caption{Optimal number of compact blocks.}
    	\label{tab-num}
    \end{table}

    \paragraph{How many blocks are replaced with video shuffle block?}
    As discussed above, the last block of ResNet layer is replaced by our compact residual block. We conduct experiments to verify whether our model can capture temporal information more effectively using more video shuffle blocks. Since a video shuffle block could be place at arbitrary ResNet layer, e.g. $res_2$, we average scores achieved by models whose number of video shuffle blocks is same. The results are reported in Table~\ref{tab-num}. It is obvious that increasing the number leads to better accuracy. Each ResNet layer having one video shuffle block (totally four) performs best.

    \begin{table}
    	\begin{center}
    	\scriptsize
    		\begin{tabular}{lll}
    			\toprule
    			Model  & Kinetics & Sth-V1 \\ 
    			\midrule
    			baseline & 71.5 & 20.2 \\
    			+ temporal shift & $\text{72.4}^\text{+0.9}$ & $\text{43.4}^\text{+23.2}$ \\
    			+ video shuffle & $\text{73.3}^\text{+1.8}$ & $\text{46.0}^\text{+25.8}$ \\
    			+ combination & $\textbf{73.5}^\textbf{+2.0}$ & $\textbf{46.6}^\textbf{+26.4}$ \\
    			\toprule
    		\end{tabular}
    	\end{center}
    	\caption{Comparison of video shuffle and temporal shift.}
    	\label{tab-comb}\vspace{-4mm}
    \end{table}

    \paragraph{Comparison of temporal shift and video shuffle.}
    Table~\ref{tab-comb} presents the respective temporal modeling ability of temporal shift and video shuffle, as well as their combined version. In comparison to temporal shift module, video shuffle is more competitive on both Kinetics and Sth-V1. Furthermore, combining these two component gains higher scores and shows superior temporal modeling capability.

    \section{Conclusion}

    In this paper, we introduced the video shuffle network, an efficient video recognition model that can conveniently learn spatio-temporal representation by inserting video shuffle into 2D CNNs. VSN not only enables 2D convolutions performing temporal modeling, but also hardly increases the overall latency. In experiments, VSN outperforms its counterparts by a great margin and further achieves state-of-the-art performance on Something-Something-V1, Something-Something-V2 and Moments in Time. We hope that our insights will inspire new efficient network designs concentrating computation and accuracy trade-off in video recognition.
	
	{\small
		\bibliographystyle{ieee}
		\bibliography{egbib}
	}
	
\end{document}